\documentclass{bmvc2k}

\usepackage{float}

\title{Sparse Autoencoders for Expert-Assisted Interpretation of Histopathology Foundation Model Embeddings}

\addauthor{\v{S}imon \v{S}m\'{i}da}{simon.smida@fit.cvut.cz}{1}
\addauthor{Wanda Benesova}{vanda.benesova@fit.cvut.cz}{1}

\addinstitution{
Czech Technical University\\
Prague, Czech Republic
}

\runninghead{\v{S}m\'{i}da, Benesova}{SAEs for Histopathology Foundation Models}

\def\etal{\emph{et al}\bmvaOneDot}

\usepackage{amsfonts}
\usepackage{booktabs}
\usepackage{microtype}
\usepackage{placeins}

\begin{document}

\maketitle

\begin{abstract}
Histopathology foundation models learn rich embedding representations, but their internal structure remains difficult to inspect. Concept-level interpretability is particularly relevant in computational pathology because pathologists reason through recurring morphological and tissue-level patterns, including tissue architecture, cellular morphology, staining variation, and artifacts. In this work, we investigate whether sparse autoencoders (SAEs) can decompose embeddings from histopathology foundation models into sparse latent features suitable for expert-assisted interpretation. We train vanilla and BatchTopK SAE variants on NCT-CRC embeddings, using CONCH as the primary foundation model and UNI2-h as an additional robustness evaluation, and characterize the resulting latent features through reconstruction fidelity, sparsity, coherence, label enrichment, and downstream class-signal preservation. The learned sparse representations preserve most downstream tissue-class information, and their latent features retrieve more coherent and label-concentrated patch groups than raw embedding dimensions or PCA components. Discussions with pathology experts further motivated a WSI-level latent visualization workflow that places feature activations into larger tissue context. Overall, the additional UNI2-h evaluation suggests that the observed behavior is not specific to CONCH, supporting the use of sparse autoencoders for expert-assisted analysis of histopathology foundation-model embeddings, while emphasizing that latent features should be treated as candidate concepts requiring pathological assessment rather than automatically validated histopathological concepts.
\end{abstract}

\section{Introduction}
\label{sec:intro}
Foundation models pretrained on large histopathology data collections are increasingly used as general-purpose backbones in computational pathology~\cite{lu2024conch, chen2024uni, vorontsov2024virchow}. Instead of training task-specific representations from scratch, many modern pipelines first map tissue patches into dense embedding spaces using pretrained models such as CONCH~\cite{lu2024conch}, UNI~\cite{chen2024uni}, or Virchow~\cite{vorontsov2024virchow}, and then reuse these embeddings for downstream analysis tasks including classification, retrieval, clustering, slide-level aggregation, and multimodal alignment. These embeddings capture variation in morphology, spatial context, staining, and other tissue properties beyond a single task label. However, they are difficult to inspect directly: individual embedding dimensions are not designed to correspond to human-understandable histopathological concepts, and relevant information may be distributed across entangled directions. This makes it difficult to determine which histopathological structures are encoded in the representation space.

Understanding what histopathology foundation models encode, how this information is organized, and how it relates to downstream model behavior remains an open challenge. Existing interpretability methods in computational pathology often focus on localization, using saliency maps, attention visualization, or multiple-instance learning heatmaps to highlight image regions associated with a prediction~\cite{selvaraju2017grad, ilse2018attention}. Such methods are useful for asking \emph{where} the evidence may be located in the input, but they provide limited insight into \emph{what} reusable histological structure is encoded in the representation space itself~\cite{poeta2023concept}.

Pathologists reason through recurring semantic abstractions such as tissue organization, cellular morphology, immune infiltration, stromal structure, staining variation, and artifacts~\cite{madabhushi2016image}. Explanations expressed at this level are often more clinically meaningful than explanations based solely on pixels, attention maps, or individual embedding dimensions. This motivates concept-based interpretability methods~\cite{kim2018tcav,koh2020cbm}, which aim to relate model behavior to higher-level human-understandable concepts. However, pathology foundation-model embeddings are learned without explicit concept supervision and may encode mixed or entangled visual structure. We therefore require representation-level methods that can identify interpretable structure directly from the embedding space.

Sparse autoencoders (SAEs) learn overcomplete sparse latent representations that reconstruct dense activation vectors while encouraging only a small subset of latent features to activate for each sample~\cite{bricken2023monosemanticity,bussmann2024batchtopk}. Unlike classical linear decompositions such as PCA, which produce dense orthogonal directions, SAEs enforce sparse feature activations and may therefore yield more selective and interpretable representation features. SAEs have recently become an important tool in mechanistic interpretability for studying sparse feature structure and polysemanticity in large neural networks~\cite{bricken2023monosemanticity}, with emerging applications in vision and computational pathology~\cite{tamura2025patch,le2024learning}. However, systematic studies of SAEs trained directly on histopathology foundation-model embeddings remain limited.

In this work, we study how sparse autoencoders can support expert-assisted interpretation of histopathology foundation-model embeddings. We train vanilla and BatchTopK~\cite{bussmann2024batchtopk} SAEs on NCT-CRC embeddings~\cite{kather2019predicting}, using CONCH~\cite{lu2024conch} as the primary model and UNI2-h~\cite{chen2024uni} to evaluate robustness to a second foundation-model embedding space. We analyze the resulting sparse latent features through quantitative characterization, spatial visualization, and downstream class-signal preservation.

We compare SAE latents against raw embedding dimensions, PCA decompositions, and random directions to evaluate whether sparse dictionary learning produces a more concentrated and coherent representation basis. Discussions with pathology experts additionally revealed that isolated top-activating patches are often insufficient for reliable interpretation of latent features. This motivated the development of a WSI-level latent visualization workflow that enables sparse features to be inspected directly within larger spatial tissue context.

Our experiments show that BatchTopK SAEs provide the strongest reconstruction--sparsity tradeoff in this setting while preserving most downstream tissue-class information. Compared with raw embedding dimensions and PCA decompositions, SAE latent features exhibit higher top-patch coherence and stronger label concentration, suggesting a more selective and coherent representation structure. Rather than treating sparse latent features as automatically validated histopathological concepts, we position them as candidate concepts that can be quantitatively characterized, spatially visualized, and assessed by pathology experts.

\begin{figure}[!htbp]
    \centering
    \includegraphics[width=\textwidth]{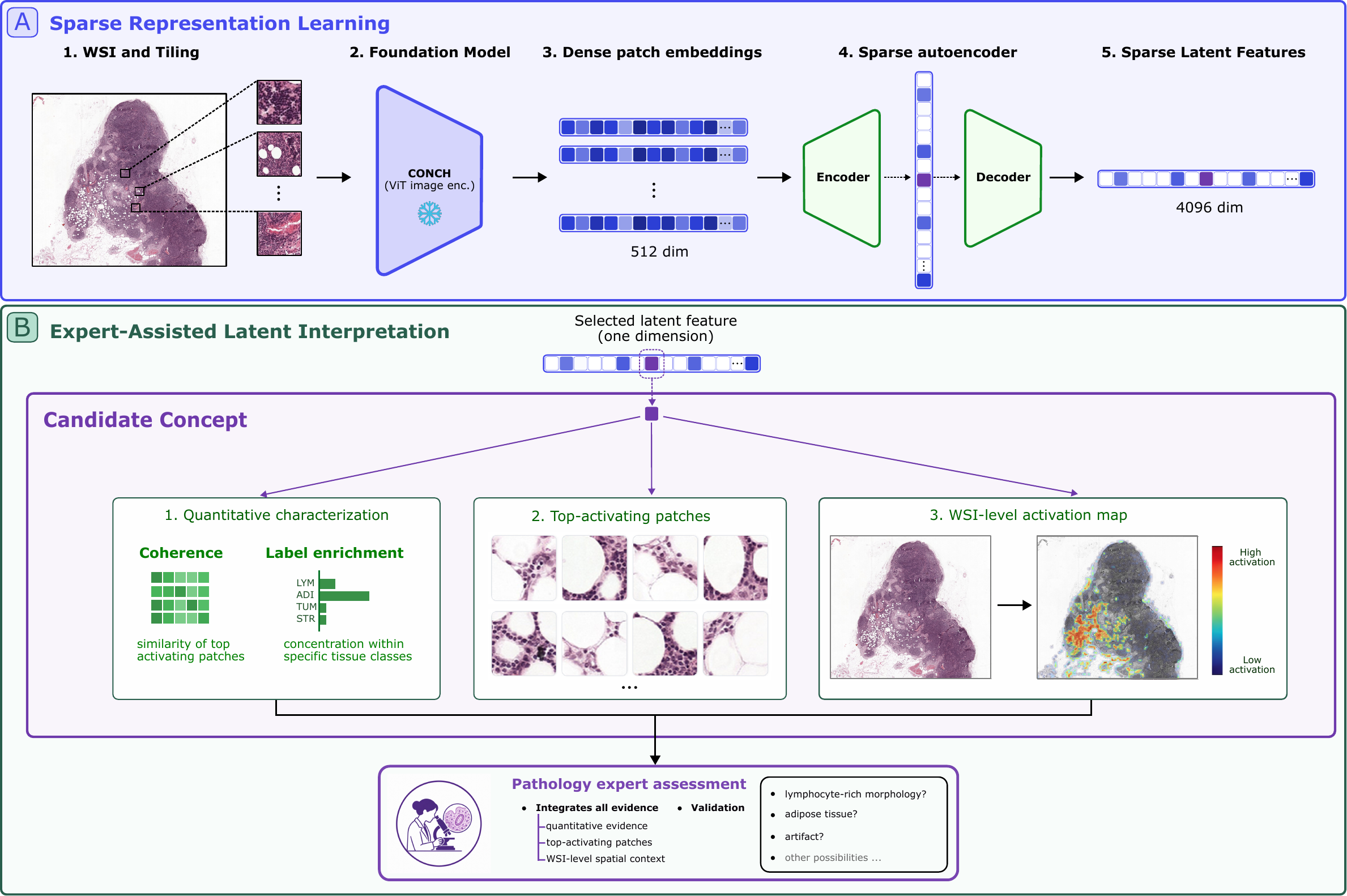}
    \caption{\textbf{Expert-assisted interpretation of histopathology foundation-model embeddings with sparse autoencoders.}
    \textbf{(A)}~\emph{Sparse representation learning.} Dense patch embeddings are decomposed into sparse latent features using a sparse autoencoder.
    \textbf{(B)}~\emph{Expert-assisted latent interpretation.} Individual latent features are analyzed through complementary forms of evidence, including quantitative characterization, top-activating patches, and WSI-level activation maps. Together these provide evidence for a candidate concept, which can then be assessed and validated by pathology experts.}
    \label{fig:pipeline_overview}
\end{figure}

\noindent
The main contributions of this work are:

\begin{itemize}
    \item We propose an expert-assisted workflow for interpreting histopathology foundation-model embeddings using sparse autoencoders.
    \item We introduce a quantitative framework for assessing SAE latent features that may correspond to candidate histopathological concepts.
    \item We develop a WSI-level latent visualization workflow motivated by pathology expert inspection requirements, enabling latent features to be inspected directly within whole-slide tissue context.
    \item We demonstrate that sparse autoencoders trained on CONCH embeddings recover coherent and label-enriched latent structures while preserving most downstream tissue-class information.
\end{itemize}

\noindent
Figure~\ref{fig:pipeline_overview} illustrates the proposed sparse-feature analysis pipeline.

\section{Related Work}
\subsection{Histopathology Foundation Models}
Foundation models have shifted computational pathology from training task-specific feature extractors toward reusing large pretrained representation backbones across diverse downstream tasks. In this paradigm, tissue patches are first encoded into dense embedding vectors using a pretrained model and these embeddings are then reused for classification, retrieval, clustering, slide-level aggregation, and multimodal analysis.

Existing histopathology foundation models differ primarily in their pretraining objectives. Vision-only models, such as UNI~\cite{chen2024uni} and Virchow~\cite{vorontsov2024virchow}, are typically trained on large pathology image collections using self-supervised objectives similar to DINO-style distillation~\cite{caron2021dino} or masked image modeling~\cite{he2022masked}. These models learn general-purpose visual representations of tissue morphology. A complementary family consists of vision-language models, which jointly learn image and text representations from pathology image--caption supervision~\cite{radford2021clip}. CONCH belongs to this family and produces semantically enriched embeddings aligned with textual pathology descriptions. Recent work suggests that vision-language pretraining may encourage representations that align more closely with semantic textual descriptions~\cite{rao2024discover,kaushik2026decomposing}. This motivates our use of CONCH as the primary model for sparse representation analysis, while UNI2-h provides a complementary vision-only embedding space for robustness evaluation.

Despite strong empirical performance, the internal organization of these embe\-dding spa\-ces remains insufficiently understood. A dense embedding may contain information about tissue type, morphology, staining variation, artifacts, and diagnostic context, but these sources of variation are not exposed as directly interpretable components. Addressing this gap is especially important in pathology, where model behavior should ideally be related back to recognizable histological patterns and expert reasoning. Our work therefore treats foundation-model embeddings themselves as the object of analysis and asks whether their structure can be decomposed into sparse and inspectable latent features.

\subsection{Interpretability in Computational Pathology}

Interpretability methods in medical imaging are often motivated by the need to relate model predictions to clinically recognizable evidence~\cite{lu2021data, diao2021human}. Common approaches provide spatial explanations through saliency maps, class-activation maps, attention visualizations, and multiple-instance learning heatmaps~\cite{selvaraju2017grad,ilse2018attention, lu2021data}. In computational pathology, these methods help localize tissue regions associated with a prediction, but provide limited insight into how reusable histological structure is organized within the representation space itself~\cite{poeta2023concept}.

Concept-based interpretability methods instead explain model behavior through higher-level abstractions rather than pixels or individual embedding dimensions. Methods such as TCAV~\cite{kim2018tcav} and concept bottleneck models~\cite{koh2020cbm} relate predictions to user-defined or annotated concepts. This level of abstraction is particularly natural in histopathology, where pathologists reason through recurring morphological and tissue-level patterns such as glandular organization, immune infiltration, stromal structure, staining variation, and artifacts. However, these approaches generally require concepts to be specified in advance together with associated annotations or examples.

This requirement becomes restrictive for histopathology foundation models. Their embedding spaces may contain diagnostic, morphological, technical, and dataset-specific structure that is not organized into predefined concept dimensions. This motivates representation-level interpretability methods that analyze embedding spaces directly, providing a natural setting for sparse representation learning.

\subsection{Sparse Autoencoders and Histopathology Representation Analysis}

Sparse autoencoders have emerged as an important tool in mechanistic interpretability for analyzing internal representations of large neural networks~\cite{bricken2023monosemanticity,cunningham2023sparse,bussmann2024batchtopk}. Their use is motivated by the observation that meaningful features may be distributed across multiple neurons or embedding dimensions, leading individual units to exhibit polysemantic behavior~\cite{elhage2022toy, bricken2023monosemanticity}. Sparse dictionary learning addresses this by representing activations as sparse combinations of learned latent features, which may provide a more interpretable basis for representation analysis.

Unlike classical linear decompositions such as PCA, sparse autoencoders learn non-orthogonal sparse feature dictionaries that may better separate reusable structure within dense embedding spaces. Recent work in language and vision-language models has shown that SAE latent features can correspond to interpretable patterns, behaviors, or semantic structures encoded within model representations~\cite{bricken2023monosemanticity,rao2024discover}. Several SAE architecture variants have subsequently been proposed to improve feature quality, sparsity control, and training stability~\cite{rajamanoharan2024improving, rajamanoharan2024jumping, gao2025scaling, bussmann2024batchtopk}.

More recently, sparse autoencoders have begun to be explored in computational pathology and medical imaging. Le \etal~\cite{le2024learning} analyzed sparse features learned from pathology foundation-model embeddings and reported biologically meaningful latent structure associated with tissue morphology, cell populations, staining variation, and artifacts. Tamura \etal~\cite{tamura2025patch} explored sparse and weakly supervised feature selection approaches for identifying pathology-relevant visual phenotypes. These studies suggest that sparse representation learning may provide a useful framework for analyzing the internal organization of histopathology foundation-model embeddings.

While prior pathology SAE studies primarily focused on biological relevance analysis and sparse feature characterization~\cite{le2024interpretability, le2024learning, le2025evaluating, tamura2025patch}, our work emphasizes expert-assisted interpretation workflows in computational pathology. In particular, we frame SAE latents as candidate concepts and introduce a WSI-level latent inspection workflow motivated by the need for larger tissue architecture and spatial context during pathologist inspection. Concurrent work has also combined pathology SAEs with expert interpretation, spatial context and molecular readouts~\cite{zhao2026compositional,kim2026picasso}.

Our UNI2-h experiments are therefore intended as a robustness check for the proposed workflow, not as a comprehensive comparison of pathology foundation-model architectures.

\section{Method}

\subsection{Terminology and Concept Framing}

We distinguish between three levels of interpretation. A \emph{latent feature} denotes one sparse dimension learned by the SAE. A \emph{candidate concept} denotes a latent feature selected for interpretation because it is supported by quantitative, visual, or spatial evidence, such as label enrichment, coherent top-activating patches, or meaningful WSI-level activation patterns. A \emph{validated concept} denotes a candidate concept that has been semantically assessed by pathologists.

This distinction is important because sparse latent features are not guaranteed to correspond to human-understandable concepts. They may be dataset-specific, artifact-associated, polysemantic, or only partially aligned with expert-defined histological abstractions.

\subsection{Dataset and Embeddings}

We train and evaluate sparse autoencoders on histopathology patch embeddings extracted from the NCT-CRC-HE-100K dataset~\cite{kather2019predicting}. The dataset contains hematoxylin--eosin (H\&E) stained colorectal cancer tissue patches of size $224\times224$ pixels organized into nine tissue classes: adipose tissue (\texttt{ADI}), background (\texttt{BACK}), debris (\texttt{DEB}), lymphocytes (\texttt{LYM}), mucus (\texttt{MUC}), smooth muscle (\texttt{MUS}), normal colon mucosa (\texttt{NORM}), cancer-associated stroma (\texttt{STR}), and colorectal adenocarcinoma epithelium (\texttt{TUM}). We use the provided training split of 100,000 patches and validation split of 7,180 patches.

For each patch, we extract embeddings using CONCH~\cite{lu2024conch} and UNI2-h~\cite{chen2024uni}. We use patch-level image embeddings as the object of analysis throughout this work. CONCH produces 512-dimensional embeddings, whereas UNI2-h produces 1536-dimensional embeddings. We refer to these vectors as \emph{raw embeddings} to indicate that they are used in their extracted, non-$\ell_2$-normalized form. This yields paired CONCH and UNI2-h embeddings for 100,000 training patches and 7,180 validation patches. Each embedding is paired with the corresponding NCT-CRC tissue label, which is used only for post-hoc analysis of latent-feature enrichment, coherence, and downstream class-signal preservation.

\subsection{Sparse Autoencoder Training}

We train sparse autoencoders to reconstruct dense patch embeddings through a sparse intermediate representation. Let $x \in \mathbb{R}^{d}$ denote a raw embedding, where $d$ is the embedding dimension of the corresponding foundation model. The SAE first centers the input using a learned bias term $b_{\mathrm{pre}}$ and computes nonnegative encoder activations
\begin{equation}
    h = \mathrm{ReLU}(W_{\mathrm{enc}}(x - b_{\mathrm{pre}}) + b_{\mathrm{enc}}),
\end{equation}
where $h \in \mathbb{R}^{m}$ and $m$ is the dictionary width. A sparse code $z$ is then obtained from $h$, and the input embedding is reconstructed as
\begin{equation}
    \hat{x} = W_{\mathrm{dec}} z + b_{\mathrm{pre}}.
\end{equation}
The learned centering term and decoder-column normalization follow standard SAE practice and prevent arbitrary scaling between latent activations and decoder directions~\cite{bricken2023monosemanticity}.

We compare two sparsity mechanisms. The vanilla SAE sets $z=h$ and encourages sparse activations with an $L_1$ penalty:
\begin{equation}
    \mathcal{L}_{\mathrm{vanilla}}
    =
    \|x-\hat{x}\|_2^2 + \lambda \|z\|_1.
\end{equation}
We sweep $\lambda \in \{10^{-2}, 3 \times 10^{-3}, 10^{-3}, 3 \times 10^{-4}\}$.

The BatchTopK SAE instead enforces sparsity by retaining only the largest activations within each training batch~\cite{bussmann2024batchtopk}. For a batch of $B$ embeddings and target average sparsity $L_0$, the method keeps the largest $B L_0$ activations across the batch and sets all others to zero. This directly controls the expected number of active latent features per embedding. At inference time, we use the calibrated global threshold obtained from the training set, following the BatchTopK procedure.

For each foundation model, SAEs are trained independently on the corresponding raw, non-$\ell_2$-normalized embeddings using mean squared reconstruction loss. For CONCH, we evaluate dictionary widths $m \in \{1024,2048,4096\}$ and BatchTopK target sparsities $L_0 \in \{8,16,32,64\}$. For UNI2-h, we use a matched 8$\times$ expansion setting: dictionary width $m=12288$ for the 1536-dimensional embeddings and target $L_0=192$, matching the target activation density of the CONCH configuration ($m=4096$, $L_0=64$). Optimization uses Adam~\cite{kingma2014adam} with learning rate $10^{-3}$, batch size 4096, and early stopping based on validation normalized reconstruction error. Model selection is based on the validation reconstruction--sparsity frontier (Figure~\ref{fig:sae-comparison}).

\subsection{Evaluation Protocol (qualitative and quantitative)}

We evaluate the learned sparse autoencoders along two complementary axes. First, we measure reconstruction quality and sparsity to determine whether the sparse latent representation faithfully preserves information from the original embedding space. Second, we analyze latent-feature behavior to test whether individual sparse features exhibit coherent and interpretable structure consistent with candidate concepts.

\paragraph{Reconstruction fidelity and sparsity.}
Let $x_i$ denote an original embedding and $\hat{x}_i$ its SAE reconstruction. We evaluate reconstruction quality using normalized mean squared error (NMSE),
\begin{equation}
    \mathrm{NMSE}
    =
    \frac{\sum_i \|x_i-\hat{x}_i\|_2^2}
         {\sum_i \|x_i\|_2^2},
\end{equation}

We additionally report explained variance and mean cosine similarity between original and reconstructed embeddings. Sparsity is characterized using the average $L_0$ value, defined as the mean number of active latent features per embedding.

\paragraph{Activation statistics and label enrichment.}
To characterize latent-feature utilization, we compute activation frequency, mean activation magnitude, and total activation mass for each feature. For each feature, we aggregate activation mass across tissue classes and compute the resulting label distribution. Class selectivity is summarized using normalized label entropy,
\begin{equation}
    H_{\mathrm{norm}}(f)
    =
    -
    \frac{1}{\log C}
    \sum_{c=1}^{C}
    p_f(c)\log p_f(c),
\end{equation}
where $p_f(c)$ denotes the fraction of activation mass assigned to class $c$ and $C$ is the number of tissue classes. Lower entropy indicates stronger concentration of activations within a small subset of tissue classes.

\paragraph{Top-activating patches.}
For each latent feature, we rank patches from the evaluated NCT-CRC patch pool by activation value and inspect the highest-activating examples. These top-patch grids provide a local visual summary of the image patterns associated with a feature. Unless otherwise stated, quantitative retrieval analyses use the top 32 patches per feature, and features with fewer than 32 active patches are excluded from top-patch coherence analyses. For qualitative figures, we show smaller representative subsets of the retrieved patches for readability.

\paragraph{Top-patch coherence and baseline comparisons.}
We quantitatively compare SAE latent features against alternative directions in the embedding space, including raw embedding dimensions, PCA components, and random directions. For each direction, we retrieve the corresponding top-activating patch sets and compute mean pairwise cosine similarity in the corresponding original embedding space together with normalized tissue-label entropy. Higher cosine similarity indicates that retrieved patches are geometrically close in the original representation space, while lower entropy indicates stronger tissue-label concentration. This analysis tests whether SAE latent features recover more coherent and semantically concentrated patch groups than alternative linear directions through the same embedding space.

\paragraph{Downstream class-signal preservation.}
We evaluate whether sparse latent representations preserve downstream tissue-class information from the original foundation-model embeddings. Linear probes are trained on the training split and evaluated on the validation split using multiple representations, including raw foundation-model embeddings, SAE reconstructions, full SAE latent codes, selected latent subsets, PCA components, and random feature subsets. For selected-feature analyses, features are ranked on the training split by their association with each tissue class, and the top-$k$ features per class are combined into a union before probe training. We report accuracy, balanced accuracy, macro-F1, macro-AUROC, and log loss. This experiment is not intended to demonstrate improved classification performance over the original foundation-model embeddings, but rather to test whether sparse latent codes preserve and concentrate class-relevant information within a smaller set of interpretable latent features.

\paragraph{Whole-slide visualization data.}
Because NCT-CRC-HE-100K contains only isolated image patches without spatial slide coordinates, WSI-level analyses were performed on five whole-slide images from the TCGA Colon Adenocarcinoma (TCGA-COAD) cohort~\cite{cancer2012comprehensive}. To match the NCT-CRC resolution of approximately 0.5\,$\mu$m/pixel, tissue regions were tiled into non-overlapping 444--456-pixel regions and resized to $224 \times 224$ patches, yielding 49,125 tissue patches (Table~\ref{tab:tcga_wsi_protocol}). For WSI-level visualization, we use the CONCH-based SAE because CONCH remains the primary model in our expert-inspection workflow. CONCH embeddings were extracted for each patch and processed using the trained SAE. Latent activations were then projected back to their original slide coordinates to generate WSI activation overlays.

\subsection{Expert-Assisted Candidate Concept Assessment}

Top-activating patch grids provide evidence about latent-feature behavior, but pathology experts indicated that isolated patches were difficult to interpret without broader tissue context. We therefore assess candidate concepts using complementary evidence: quantitative characterization, top-activating patches, and WSI-level activation maps (Figure~\ref{fig:candidate-concepts}).

While top-activating patches reveal image structures associated with a latent feature, WSI-level activation maps reveal how the same feature deploys within larger tissue architecture. This additional context can help distinguish meaningful tissue patterns from artifact-associated or fragmented activation behavior.

To support expert inspection, we developed an interactive interface linking latent features, top-activating patches, and WSI-level activation maps (Appendix~\ref{app:qualitative}). In our framework, latent features are treated as candidate concepts whose interpretation is informed by multiple forms of evidence rather than patch-level retrieval alone.

\section{Results}
We organize the results around the primary CONCH analysis, followed by a compact UNI2-h robustness evaluation. First, we compare SAE variants along the re\-construction--sparsity tradeoff. We then characterize active latent features and compare their retrieval behavior against embedding-space baselines. Finally, we examine how patch-level retrieval and WSI-level visualization support expert interpretation.

\subsection{Primary CONCH analysis: reconstruction--sparsity tradeoff}
\begin{figure}[!htbp]
    \centering
    \includegraphics[width=0.8\textwidth]{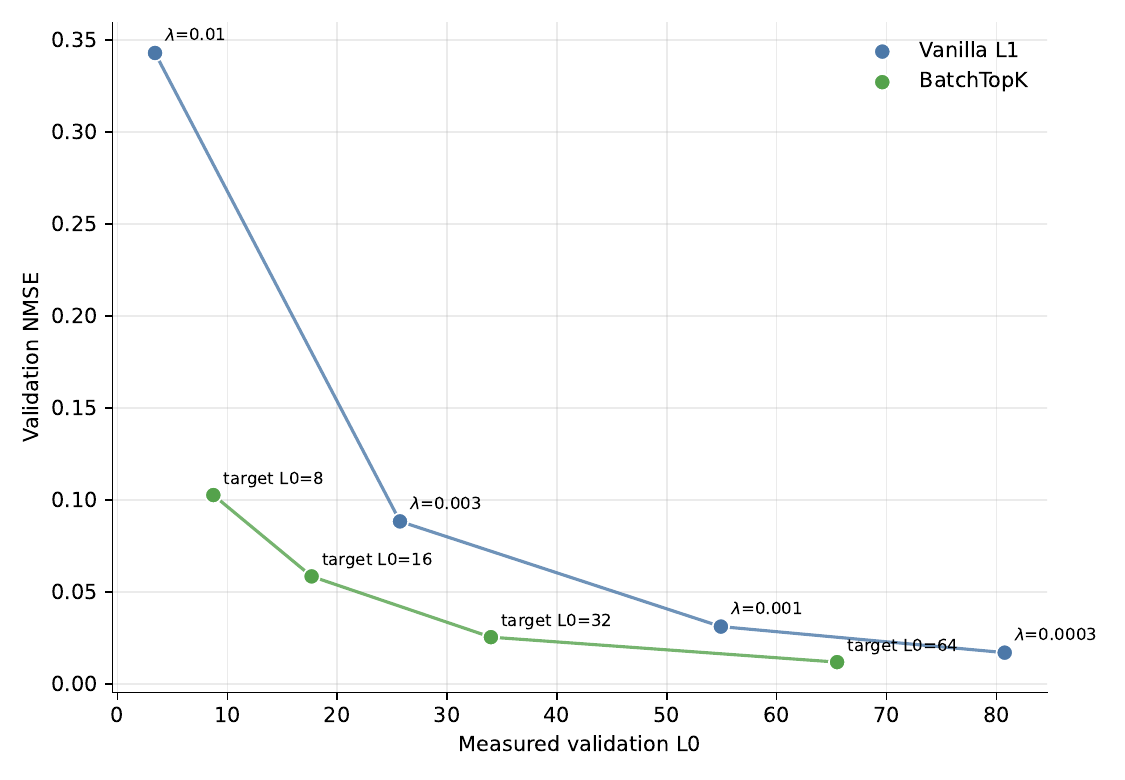}
    \caption{\textbf{Reconstruction--sparsity tradeoff for evaluated SAE variants.} BatchTopK SAEs form a consistently better reconstruction--sparsity frontier than vanilla $L_1$ SAEs across evaluated dictionary widths and sparsity settings. Increasing target sparsity $L_0$ improves reconstruction fidelity while preserving sparse latent structure.}
    \label{fig:sae-comparison}
\end{figure}

We first compare selected SAE variants by their reconstruction--sparsity tradeoff (Figure~\ref{fig:sae-comparison}). For dictionary width $m=4096$, BatchTopK models form a clear frontier as the target sparsity increases from $L_0=8$ to $L_0=64$.  BatchTopK SAEs also outperform vanilla $L_1$ SAEs at comparable sparsity levels. In this seed-0 sweep, the best-performing BatchTopK model ($m=4096$, $L_0=64$) reaches validation NMSE of 0.0117 at measured $L_0=65.50$, whereas the closest high-fidelity vanilla SAE reaches NMSE of 0.0169 despite using a denser representation with measured $L_0=80.76$. This model additionally achieves explained variance 0.978 and reconstruction cosine similarity 0.994. We therefore use this BatchTopK configuration for subsequent latent-feature analyses.

\subsection{High reconstruction stability and low dictionary utilization}

We evaluate reconstruction stability across five seeds. The selected BatchTopK configuration ($m=4096$, $L_0=64$) achieves validation NMSE $0.0117 \pm 0.0001$, while the $L_0=32$ model achieves $0.0252 \pm 0.0003$. Subsequent analyses use the $m=4096$, $L_0=64$ model (seed 1).

\begin{figure}[!htbp]
    \centering
    \includegraphics[width=\textwidth]{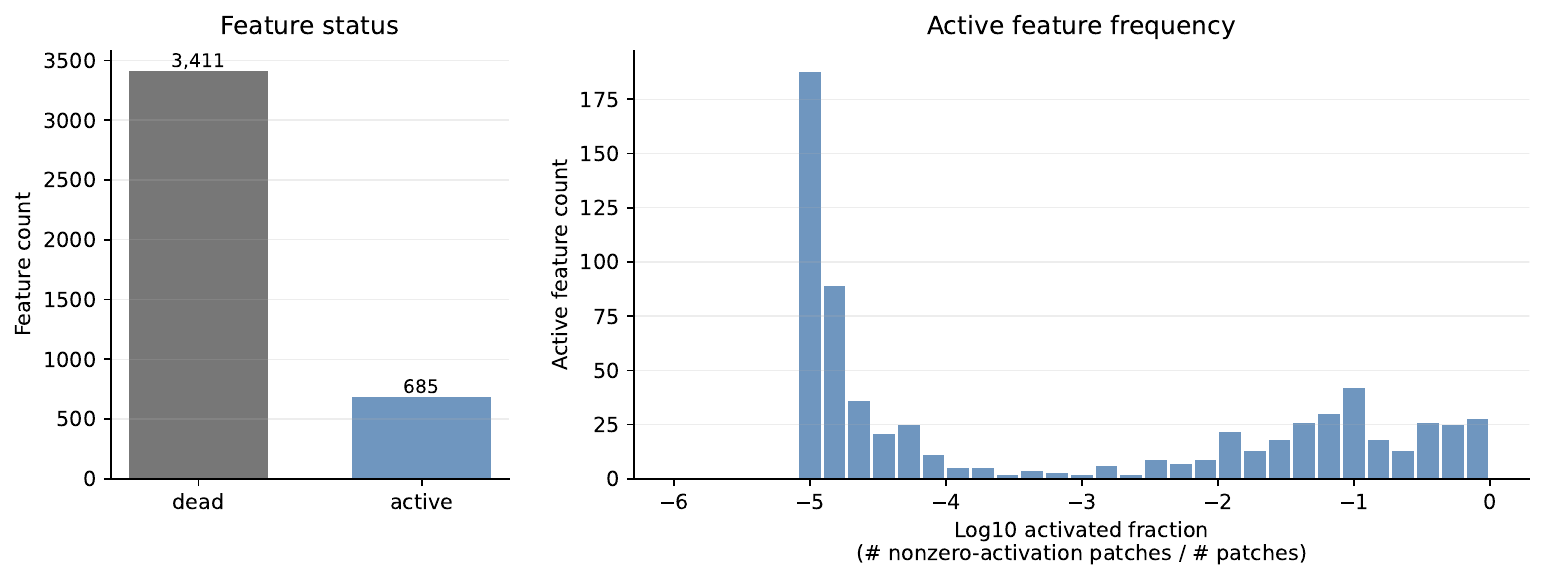}
    \caption{\textbf{SAE dictionary utilization.}
    Although the selected BatchTopK SAE reconstructs CONCH embeddings with high fidelity, only a minority of dictionary elements activate on the NCT-CRC dataset. The figure summarizes alive and dead latent features (\textit{left}) together with the activation-frequency distribution of active features (\textit{right}).}
    \label{fig:sae-dictionary}
\end{figure}

Dictionary utilization remains low (Figure~\ref{fig:sae-dictionary}). In the selected model, only 685 of 4096 latent features (16.7\%) activate at least once across the combined train and validation split; on the validation split alone, the alive-feature fraction is $0.131 \pm 0.005$ across the five $L_0=64$ seeds. Thus, accurate reconstruction on NCT-CRC uses only a small subset of the learned dictionary.

A width ablation reveals a tradeoff between reconstruction and utilization. At $L_0=64$, dictionary widths $m=1024$, $2048$, and $4096$ achieved NMSE 0.0141, 0.0124, and 0.0115, with alive-feature fractions 32.1\%, 22.0\%, and 16.7\%, respectively. Smaller dictionaries improve utilization, while the selected $m=4096$ model provides the best reconstruction. Additional model-selection and ablation results are provided in Appendices~\ref{app:training} and~\ref{app:width}.

\subsection{Active latent features are sparse, selective, and label-enriched}

We next analyze the activation behavior of the active latent subset. Activation frequencies are highly sparse, with a median activation frequency of $4.67 \times 10^{-5}$ across active features. Many latent features therefore activate only on a very small subset of image patches.

At the same time, many active features exhibit strong tissue-label enrichment despite being learned without supervision. The median label entropy across active features is 0.256 nats, corresponding to normalized entropy 0.116. This indicates that activation mass is often concentrated within a small subset of tissue classes. Class-frequency normalization produced the same qualitative conclusion. Several latent features exhibit highly selective activation behavior for individual tissue categories. For example, latent 0058 assigns 98.9\% of its activation mass to lymphocyte patches, latent 0150 assigns 98.1\% to adipose tissue, latent 1883 assigns 98.1\% to tumor epithelium, and latent 1966 activates exclusively on background patches.

\begin{figure}[!htbp]
    \centering
    \includegraphics[width=\textwidth]{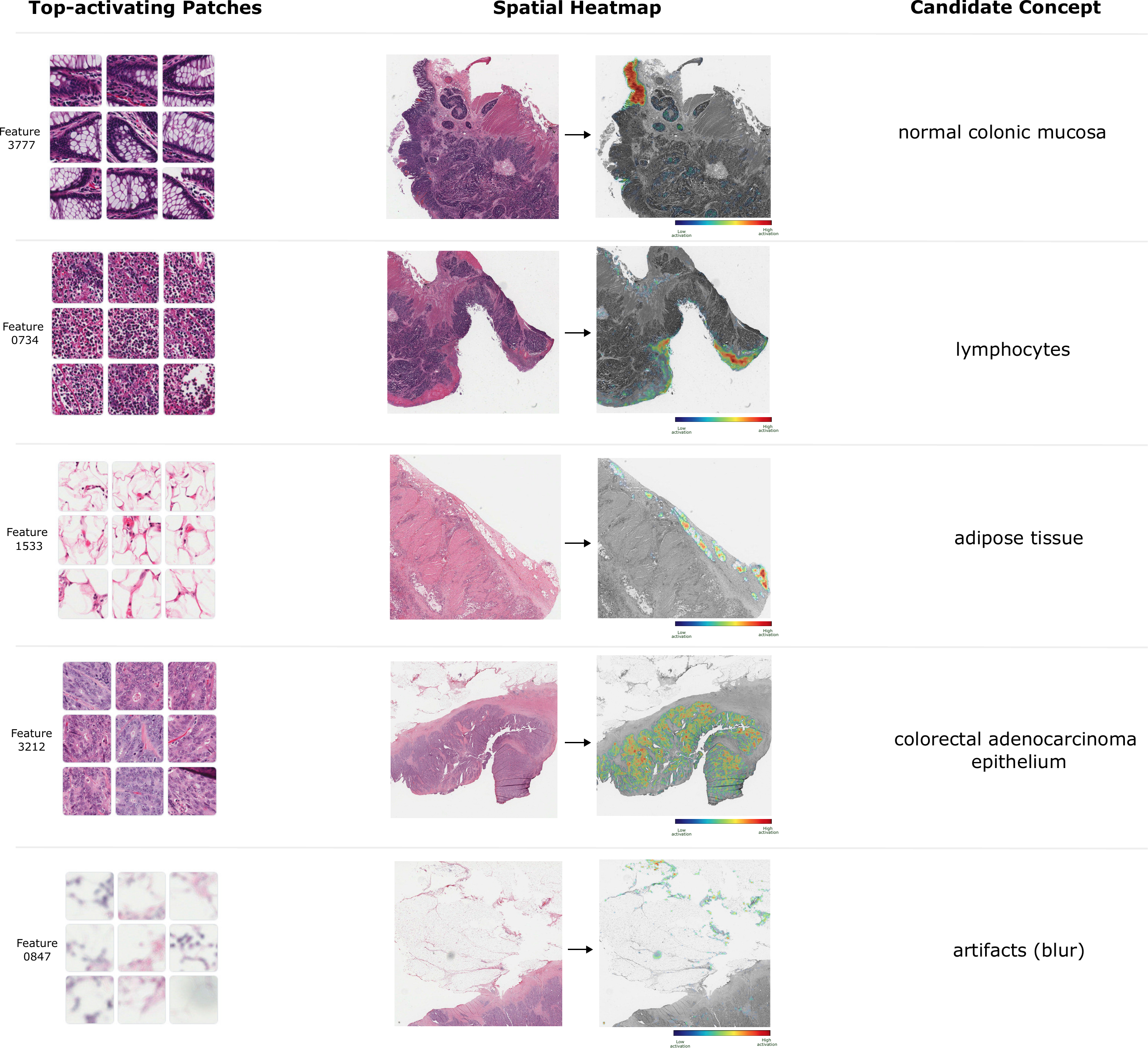}
    \caption{\textbf{Representative SAE latent features as candidate histopathological concepts.} Each row shows the highest-activating patches for a selected latent feature together with representative WSI-level deployment patterns. Many latent features exhibit visually coherent and label-enriched activation behavior, although some remain mixed or ambiguous, supporting our interpretation of SAE latents as candidate concepts rather than validated biological abstractions.}
    \label{fig:candidate-concepts}
\end{figure}

Representative active features often retrieve visually coherent histological or technical patterns, including lymphocyte-rich regions, adipose tissue, tumor-associated morphology, stromal structure, artifacts, and background (Figure~\ref{fig:candidate-concepts}). These observations suggest that at least part of the active latent features capture recurring tissue-associated structure present in the original CONCH embedding space. Additional qualitative examples, including ambiguous and artifact-associated features, are provided in Appendix~\ref{app:qualitative}.

\subsection{Latent features retrieve more coherent patch groups than alternative embedding directions}

We compare SAE latent features against raw CONCH embedding dimensions, PCA components, and random directions using top-patch retrieval analysis. For each feature or direction, we retrieve top-scoring patches and compute mean pairwise cosine similarity in the original CONCH embedding space.

\begin{figure}[!htbp]
    \centering
    \includegraphics[width=\textwidth]{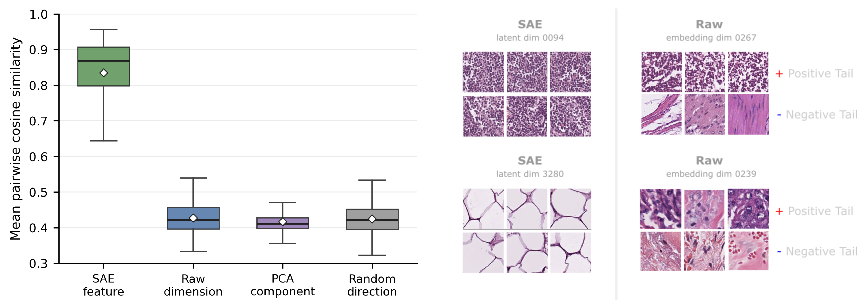}
    \caption{\textbf{SAE latent features retrieve more coherent patch groups than alternative embedding directions.}
    \textit{Left:} mean pairwise cosine similarity of retrieved patch sets in the original CONCH embedding space. Raw dimensions, PCA components, and random directions are evaluated using the union of positive and negative tails.
    \textit{Right:} retrieval examples. SAE latents retrieve coherent patch groups through a single nonnegative activation pattern, whereas raw dimensions can associate different structures with opposite tails.}
    \label{fig:sae_vs_baselines}
\end{figure}

SAE latent features retrieve more coherent patch groups than alternative embedding directions. Using the union of positive and negative tails for signed directions, SAE features achieve mean coherence 0.835, compared with 0.427 for raw dimensions, 0.417 for PCA components, and 0.425 for random directions. Figure~\ref{fig:sae_vs_baselines} summarizes the quantitative comparison and shows representative retrieval examples.

This difference reflects the asymmetry between sparse latent features and signed axes. SAE features activate through a single nonnegative activation pattern, whereas positive and negative tails of a raw embedding dimension may correspond to different visual structures. Qualitative retrieval examples show the same behavior: SAE latents produce more localized patch groups, while raw dimensions can associate unrelated patterns with opposite tails.

As a sensitivity analysis, we also evaluated a stronger best-of-tail baseline for signed directions. This narrows the gap but preserves the same ordering: SAE features achieve mean coherence 0.835, compared with 0.749 for raw dimensions, 0.473 for PCA components, and 0.748 for random directions. Additional sensitivity analyses over retrieval size and tail handling are provided in Appendix~\ref{app:tails}.

\subsection{Sparse latent representations preserve tissue-class information}\label{sec:probes}

\begin{figure}[!htbp]
    \centering
    \includegraphics[width=\textwidth]{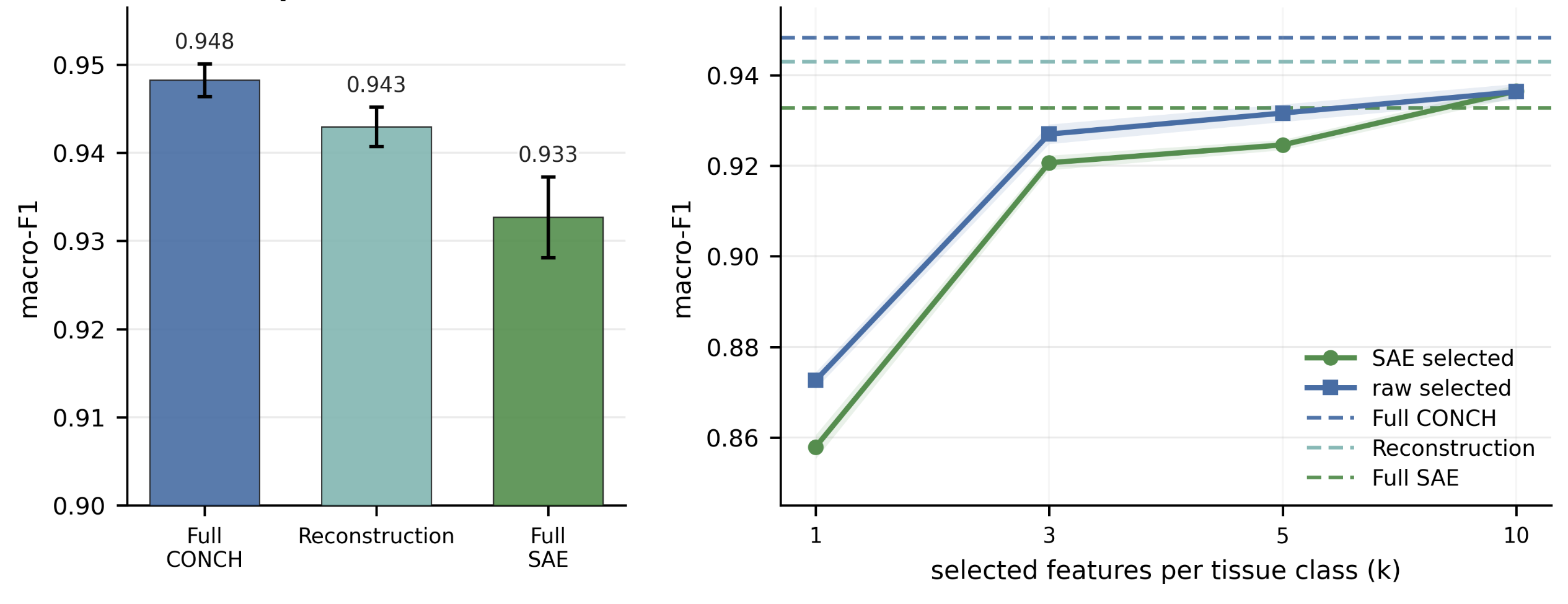}
    \caption{\textbf{Sparse latent representations preserve downstream tissue-class information.}
    \textit{Left:} Linear-probe macro-F1 for full CONCH embeddings, SAE reconstructions, and full SAE latent representations. Most tissue-class information is retained after sparse decomposition.
    \textit{Right:} Linear-probe macro-F1 as a function of the number of selected features per tissue class. Selected SAE features retain most classification performance with relatively few features and approach full latent-representation performance as additional features are included. Dashed lines indicate the full-representation baselines.}
    \label{fig:sae-classif-signal}
\end{figure}

Finally, we evaluate whether SAE representations preserve downstream tissue-class information (Figure~\ref{fig:sae-classif-signal}). A linear probe trained on full CONCH embeddings achieves validation macro-F1 0.948. SAE reconstructions preserve nearly all of this signal, reaching macro-F1 0.943, while the full sparse latent representation achieves macro-F1 0.933.

Post-hoc selected latent subsets also retain substantial class information. Using the top three selected SAE features per tissue class (27 features total), the linear probe achieves macro-F1 0.921, compared with 0.615 for random latent subsets of the same size. Increasing the selected subset to 90 SAE features improves macro-F1 to 0.936.

Raw embedding dimensions and PCA baselines remain competitive in downstream classification, so these results should not be interpreted as evidence that SAE representations improve predictive performance over the original CONCH embeddings. Instead, they demonstrate that sparse latent representations preserve and partially concentrate class-relevant structure within a more inspectable sparse basis.

\subsection{Robustness to a second foundation model}

To evaluate whether the proposed workflow depends on a particular pathology foundation model, we repeated the main quantitative analysis on UNI2-h using the same NCT-CRC train and validation splits. SAEs were trained independently for each backbone using the same training and evaluation protocol, matched expansion factor, and target activation density. Reconstruction and sparse-representation statistics are summarized in Table~\ref{tab:backbone_reconstruction}.

\begin{table}[H]
\centering
\caption{\textbf{Reconstruction and sparse-representation statistics} on NCT-CRC for CONCH and UNI2-h foundation models. Values are validation mean $\pm$ SD over five training seeds; alive fraction is measured on the validation split.}
\label{tab:backbone_reconstruction}
\scriptsize
\resizebox{\columnwidth}{!}{%
\begin{tabular}{lrrrrrr}
\hline
Backbone &
$d$ &
$m$ &
NMSE $\downarrow$ &
EV $\uparrow$ &
Measured $L_0$ &
Alive (\%)\\
\hline
CONCH
& 512
& 4,096
& $0.01169\!\pm\!0.00014$
& $0.97820\!\pm\!0.00027$
& $65.72\!\pm\!0.26$
& $13.12\!\pm\!0.49$
\\

UNI2-h
& 1,536
& 12,288
& $0.16733\!\pm\!0.00021$
& $0.79792\!\pm\!0.00026$
& $223.34\!\pm\!0.48$
& $73.28\!\pm\!0.67$
\\
\hline
\end{tabular}%
}
\end{table}

Compared with CONCH, UNI2-h achieves lower reconstruction fidelity but substantially higher dictionary utilization, indicating that the reconstruction--sparsity frontier is backbone dependent rather than identical across foundation models. Nevertheless, the main qualitative conclusions remain consistent across both backbones (Table~\ref{tab:backbone_downstream}). For UNI2-h, the 6,118 SAE features eligible for top-32 retrieval reach mean coherence 0.605, compared with 0.497 for raw dimensions, 0.485 for random directions and 0.290 for PCA components under the best-tail setting (Appendix~\ref{app:uni2}). On both foundation models, SAE latent features retrieve more coherent and label-concentrated patch groups than raw embedding dimensions, PCA components, and random directions, while sparse representations preserve most downstream tissue-class information. Together, these results suggest that the behavior of the proposed SAE-based interpretation workflow is not specific to a single pathology foundation model, although a comprehensive comparison of foundation-model architectures remains outside the scope of this work.

\begin{table}[!htbp]
\centering
\caption{\textbf{Interpretability and downstream evaluation for CONCH and UNI2-h on NCT-CRC.} Coherence and entropy summarize top-32 retrieval of SAE features (303 CONCH and 6,118 UNI2-h features), measured in each model's own embedding space, so absolute values should not be compared across models. F1 values are point estimates from a linear-probe protocol matched across both models (L-BFGS with strong-Wolfe line search); the CONCH values therefore differ slightly from the bootstrap means in Section~\ref{sec:probes}.}
\label{tab:backbone_downstream}
\small
\begin{tabular}{lrrrrr}
\hline
Backbone & Coherence$\uparrow$ & Entropy$\downarrow$ & Raw F1 & Recon. F1 & Latent F1\\
\hline
CONCH
& 0.8348
& 0.0898
& 0.9497
& 0.9448
& 0.9358
\\

UNI2-h
& 0.6052
& 0.1010
& 0.9553
& 0.9573
& 0.9542
\\
\hline
\end{tabular}
\end{table}

\noindent
Additional UNI2-h details and top-patch examples are provided in Appendix~\ref{app:uni2}.

\subsection{WSI-level deployment supports expert assessment}

WSI-level visualization provides information that is not available from patch retrieval alone (Figure~\ref{fig:candidate-concepts}). While top-activating patches reveal local image structures associated with a latent feature, WSI activation overlays reveal how the same feature appears within larger tissue architecture. During expert inspection, this additional spatial context proved valuable because isolated patches were often difficult to interpret on their own. In qualitative examples, WSI overlays provided spatial context for inspecting whether activations appeared localized, diffuse, fragmented, or artifact-associated. These observations support our framing of SAE latent features as candidate concepts whose interpretation benefits from quantitative characterization, patch-level evidence, and spatial deployment analysis.

\section{Conclusion}

We investigated sparse autoencoders as a tool for analyzing histopathology foundation-model embeddings. Using CONCH embeddings from the NCT-CRC dataset as the primary analysis, with an additional UNI2-h robustness evaluation, we showed that BatchTopK SAEs achieve high-fidelity sparse reconstruction while preserving most downstream tissue-class information. The learned latent features exhibit coherent visual behavior and strong tissue-label enrichment, and retrieve more coherent patch groups than raw embedding dimensions or PCA components. Together, these results suggest that sparse dictionary learning can make parts of histopathology foundation-model embedding spaces more inspectable.

We emphasize that SAE latent features should be viewed as candidate concepts rather than automatically validated histopathological concepts. In our framework, quantitative characterization, top-patch inspection, and WSI-level visualization provide complementary evidence that supports expert interpretation. Important limitations include the use of a single main dataset, limited foundation-model coverage, the absence of systematic pathologist evaluation, and the correlational nature of the analysis. Future work should test whether these candidate concepts are stable across datasets and foundation models, evaluate them systematically with pathologists, and examine their causal relevance through latent interventions.

\bibliography{egbib}

\clearpage
\appendix
\FloatBarrier
\section{Additional Training and Model-Selection Details}\label{app:training}

All CONCH sparse autoencoders in this section were trained on raw, non-$\ell_2$-normalized CONCH embeddings from the NCT-CRC training split. Each model used a 512-dimensional input, mean-squared reconstruction loss, Adam optimizer with learning rate $10^{-3}$, batch size 4096, and early stopping by validation normalized mean-squared error (NMSE). BatchTopK models were parameterized by target average sparsity $L_0$; vanilla $L_1$ models were parameterized by the sparsity coefficient $\lambda$.

Table~\ref{tab:model_selection_configs} reports the validation metrics for the model-selection runs with complete validation outputs. The selected model used in the main analyses is the BatchTopK SAE with $m=4096$, target $L_0=64$, and seed 1. On the validation split this model achieved NMSE $0.0115$, explained variance $0.978$, reconstruction cosine similarity $0.994$, and measured $L_0=66.03$.

\begin{table}[H]
\centering
\scriptsize
\caption{\textbf{Validation metrics for SAE model-selection runs with complete validation outputs.} BatchTopK models are specified by target $L_0$; vanilla $L_1$ models are specified by the sparsity coefficient $\lambda$. Alive fraction is measured on the NCT-CRC validation split.}
\label{tab:model_selection_configs}
\begin{tabular}{llrrrrrrr}
\toprule
SAE & Sparsity & $m$ & Seed & NMSE $\downarrow$ & EV $\uparrow$ & Cosine $\uparrow$ & measured $L_0$ & Alive \% \\
\midrule
BatchTopK & $L_0=8$ & 1024 & 0 & 0.1073 & 0.800 & 0.946 & 8.74 & 10.4 \\
BatchTopK & $L_0=64$ & 1024 & 0 & 0.0141 & 0.974 & 0.993 & 64.00 & 25.0 \\
BatchTopK & $L_0=8$ & 2048 & 0 & 0.1041 & 0.806 & 0.947 & 9.00 & 6.3 \\
BatchTopK & $L_0=16$ & 2048 & 0 & 0.0612 & 0.886 & 0.969 & 16.96 & 9.3 \\
BatchTopK & $L_0=32$ & 2048 & 0 & 0.0259 & 0.952 & 0.987 & 33.60 & 11.3 \\
BatchTopK & $L_0=64$ & 2048 & 0 & 0.0124 & 0.977 & 0.994 & 65.40 & 17.3 \\
BatchTopK & $L_0=8$ & 4096 & 0 & 0.1026 & 0.809 & 0.948 & 8.73 & 3.6 \\
BatchTopK & $L_0=16$ & 4096 & 0 & 0.0583 & 0.891 & 0.971 & 17.67 & 5.2 \\
BatchTopK & $L_0=32$ & 4096 & 0 & 0.0253 & 0.953 & 0.987 & 33.99 & 7.8 \\
BatchTopK & $L_0=64$ & 4096 & 0 & 0.0117 & 0.978 & 0.994 & 65.50 & 12.1 \\
\midrule
Vanilla $L_1$ & $\lambda=10^{-2}$ & 1024 & 0 & 0.2456 & 0.542 & 0.883 & 4.69 & 79.8 \\
Vanilla $L_1$ & $\lambda=3{\times}10^{-3}$ & 1024 & 0 & 0.0858 & 0.840 & 0.958 & 24.03 & 84.9 \\
Vanilla $L_1$ & $\lambda=10^{-2}$ & 2048 & 0 & 0.2835 & 0.472 & 0.859 & 3.81 & 48.2 \\
Vanilla $L_1$ & $\lambda=3{\times}10^{-3}$ & 2048 & 0 & 0.0852 & 0.841 & 0.958 & 25.06 & 71.8 \\
Vanilla $L_1$ & $\lambda=10^{-3}$ & 2048 & 0 & 0.0303 & 0.944 & 0.985 & 58.22 & 59.7 \\
Vanilla $L_1$ & $\lambda=3{\times}10^{-4}$ & 2048 & 0 & 0.0150 & 0.972 & 0.993 & 95.97 & 57.8 \\
Vanilla $L_1$ & $\lambda=10^{-2}$ & 4096 & 0 & 0.3430 & 0.361 & 0.823 & 3.42 & 30.4 \\
Vanilla $L_1$ & $\lambda=3{\times}10^{-3}$ & 4096 & 0 & 0.0882 & 0.836 & 0.957 & 25.72 & 52.1 \\
Vanilla $L_1$ & $\lambda=10^{-3}$ & 4096 & 0 & 0.0311 & 0.942 & 0.985 & 54.93 & 46.9 \\
Vanilla $L_1$ & $\lambda=3{\times}10^{-4}$ & 4096 & 0 & 0.0169 & 0.969 & 0.992 & 80.76 & 50.6 \\
\bottomrule
\end{tabular}
\end{table}

We additionally repeated the strongest $m=4096$ BatchTopK settings across five random seeds. Table~\ref{tab:seed_stability} shows that both target sparsity settings were stable, with the $L_0=64$ configuration providing the strongest reconstruction fidelity.

\begin{table}[H]
\centering
\small
\setlength{\tabcolsep}{4pt}
\caption{\textbf{Seed stability for BatchTopK SAEs with $m=4096$.} Values are mean $\pm$ standard deviation across five random seeds.}
\label{tab:seed_stability}
\begin{tabular}{lcccc}
\toprule
Configuration & NMSE & EV & measured $L_0$ & Val. alive fraction \\
\midrule
$m=4096,\ L_0=32$ & 0.0252 $\pm$ 0.0003 & 0.9531 $\pm$ 0.0005 & 33.93 $\pm$ 0.17 & 0.072 $\pm$ 0.005 \\
$m=4096,\ L_0=64$ & 0.0117 $\pm$ 0.0001 & 0.9782 $\pm$ 0.0003 & 65.72 $\pm$ 0.26 & 0.131 $\pm$ 0.005 \\
\bottomrule
\end{tabular}
\end{table}

\FloatBarrier
\section{Dictionary Width and Utilization Ablations}\label{app:width}

Dictionary utilization was evaluated to address the possibility that the learned dictionary contains many inactive features. At target $L_0=64$, increasing width improved reconstruction but reduced the fraction of alive features. The selected $m=4096$ dictionary therefore trades lower NMSE for lower utilization, whereas smaller dictionaries use a larger fraction of their capacity.

\begin{table}[H]
\centering
\small
\setlength{\tabcolsep}{3.5pt}
\caption{\textbf{Effect of dictionary width for BatchTopK SAEs at target $L_0=64$.} Active features are counted on the combined train and validation split. Coherence and entropy are computed from top-32 patch retrieval over eligible active features.}
\label{tab:width_ablation}
\begin{tabular}{rrrrrrr}
\toprule
$m$ & NMSE $\downarrow$ & EV $\uparrow$ & measured $L_0$ & active & coherence $\uparrow$ & entropy $\downarrow$ \\
\midrule
1024 & 0.0141 & 0.9736 & 64.00 & 329 / 1024 (32.1\%) & 0.833 & 0.101 \\
2048 & 0.0124 & 0.9770 & 65.40 & 450 / 2048 (22.0\%) & 0.828 & 0.101 \\
4096 & 0.0115 & 0.9785 & 66.03 & 685 / 4096 (16.7\%) & 0.835 & 0.090 \\
\bottomrule
\end{tabular}
\end{table}

\begin{figure}[H]
\centering
\includegraphics[width=0.6\textwidth]{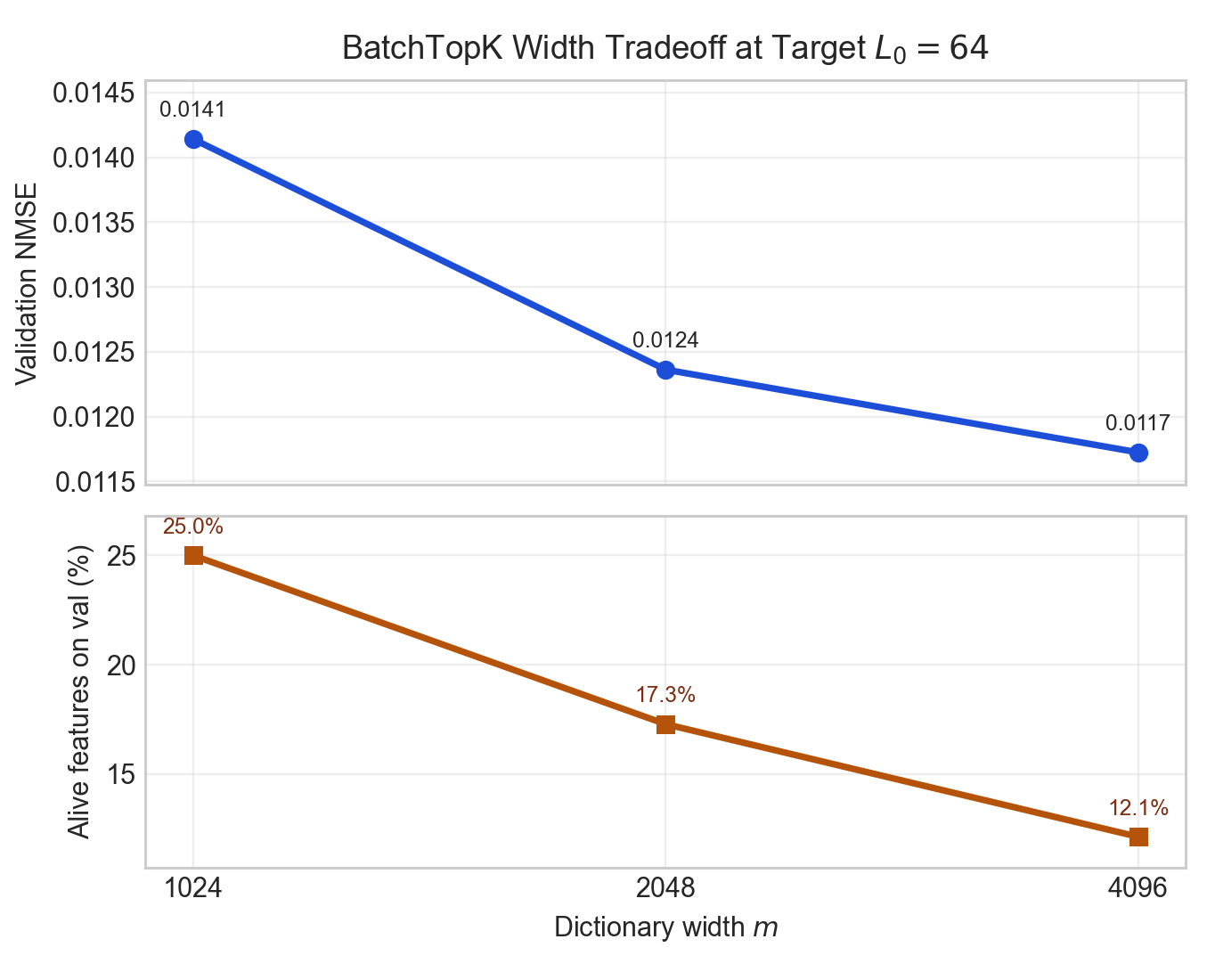}
\caption{\textbf{Dictionary-width tradeoff at target $L_0=64$.} Increasing dictionary width improves reconstruction fidelity, whereas smaller dictionaries use a larger fraction of their latent features.}
\label{fig:width_tradeoff}
\end{figure}

\FloatBarrier
\section{Label-Entropy Robustness}\label{app:entropy}

For each active feature $f$, the main analysis computes normalized label entropy from the activation-mass distribution $p_f(c)$. To check that class imbalance is not driving the conclusion, we also compute a class-frequency-normalized distribution
\begin{equation}
q_f(c) =
\frac{p_f(c)/\pi(c)}
{\sum_{c'} p_f(c')/\pi(c')},
\end{equation}
where $\pi(c)$ is the empirical class frequency over the combined NCT-CRC train and validation splits. Entropy is then computed from $q_f$ and normalized by $\log 9$.

Table~\ref{tab:entropy_robustness} shows that class-frequency normalization produces the same qualitative conclusion: many active features remain low-entropy after correcting for the NCT-CRC class distribution.

\begin{table}[H]
\centering
\small
\setlength{\tabcolsep}{4pt}
\caption{\textbf{Entropy robustness for active features of the selected SAE.} Raw entropy is measured in nats. Normalized entropy divides by $\log 9$. Class-frequency-normalized entropy first reweights each feature's activation-mass distribution by the inverse train+validation class frequency and then divides by $\log 9$.}
\label{tab:entropy_robustness}
\begin{tabular}{lccc}
\toprule
Summary over active features & Raw entropy & Normalized entropy &
\begin{tabular}[c]{@{}c@{}}Class-frequency\\normalized\end{tabular} \\
\midrule
Median & 0.256 & 0.116 & 0.122 \\
Mean & 0.550 & 0.250 & 0.252 \\
25th percentile & 0.000 & 0.000 & 0.000 \\
75th percentile & 0.807 & 0.367 & 0.381 \\
\bottomrule
\end{tabular}
\end{table}

\FloatBarrier
\section{Top-Patch Retrieval Sensitivity}\label{app:tails}

The main top-patch analysis uses the top 32 patches for each eligible feature. Table~\ref{tab:topk_sensitivity} repeats the comparison for top-16, top-32, and top-64 retrieval. SAE latent features remain more coherent and label-concentrated than raw CONCH dimensions, PCA components, and random directions under the signed-axis union comparison.

\begin{table}[H]
\centering
\caption{\textbf{Sensitivity to the number of retrieved top patches.} SAE rows use top activations of the selected BatchTopK model. Baseline rows use the union of positive and negative signed-axis tails, as in the main comparison.}
\label{tab:topk_sensitivity}
\begin{tabular}{llccc}
\toprule
Method & Metric & Top-16 & Top-32 & Top-64 \\
\midrule
SAE & Coherence & 0.842 & 0.835 & 0.822 \\
Raw dimensions & Coherence & 0.422 & 0.427 & 0.430 \\
PCA & Coherence & 0.421 & 0.417 & 0.414 \\
Random & Coherence & 0.419 & 0.425 & 0.429 \\
\midrule
SAE & Entropy & 0.081 & 0.090 & 0.103 \\
Raw dimensions & Entropy & 0.568 & 0.596 & 0.622 \\
PCA & Entropy & 0.768 & 0.819 & 0.854 \\
Random & Entropy & 0.578 & 0.608 & 0.632 \\
\bottomrule
\end{tabular}
\end{table}

Signed linear directions have two tails, whereas SAE latents are nonnegative. Table~\ref{tab:tail_handling} therefore compares the signed-union setting with a best-tail setting that gives each raw, PCA, or random direction its more coherent tail. The best-tail control narrows the gap for raw and random directions, but the SAE features retain the highest mean coherence while avoiding the ambiguity of two opposite signed tails.

\begin{table}[H]
\centering
\small
\setlength{\tabcolsep}{4pt}
\caption{\textbf{Sensitivity to signed-tail handling for top-32 retrieval.} Raw dimensions, PCA components, and random directions are signed axes. The signed-union setting combines positive and negative tails; the best-tail setting gives each signed axis the more coherent of its two tails. SAE latents are nonnegative, so the same top-activation set is used in both columns.}
\label{tab:tail_handling}
\begin{tabular}{lcccc}
\toprule
Method &
\begin{tabular}[c]{@{}c@{}}Union\\coherence $\uparrow$\end{tabular} &
\begin{tabular}[c]{@{}c@{}}Best-tail\\coherence $\uparrow$\end{tabular} &
\begin{tabular}[c]{@{}c@{}}Union\\entropy $\downarrow$\end{tabular} &
\begin{tabular}[c]{@{}c@{}}Best-tail\\entropy $\downarrow$\end{tabular} \\
\midrule
SAE feature & 0.835 & 0.835 & 0.090 & 0.090 \\
Raw CONCH dim & 0.427 & 0.749 & 0.596 & 0.219 \\
PCA component & 0.417 & 0.473 & 0.819 & 0.711 \\
Random direction & 0.425 & 0.748 & 0.608 & 0.223 \\
\bottomrule
\end{tabular}
\end{table}

\FloatBarrier
\section{Selected-Feature Linear-Probe Protocol}\label{app:probes}

Selected-feature probes were designed to avoid validation leakage. For each class $c$, we formed a one-vs-rest training label and ranked candidate dimensions by absolute Pearson correlation with that label using the training split only. For a selected value of $k$, the top $k$ dimensions for each class were combined into a union. The validation split was used only after feature selection, standardization, hyperparameter selection, and probe fitting were determined from training data. Random SAE subsets were drawn from train-alive SAE features with the same size as the selected SAE union.

Table~\ref{tab:selected_feature_protocol} reports the selected-feature validation macro-F1 for $k\in\{1,3,5,10\}$ per class. These results support the main conclusion that sparse latent codes preserve and concentrate tissue-class signal, while raw dimensions remain a strong predictive baseline.

\begin{table}[H]
\centering
\small
\setlength{\tabcolsep}{4pt}
\caption{\textbf{Train-only selected-feature probe results for the selected BatchTopK SAE.} Features are ranked by absolute Pearson correlation with one-vs-rest class labels on the training split only. The table reports validation macro-F1 mean $\pm$ standard deviation. Random subsets sample the same number of alive SAE features as the selected SAE union for each $k$; raw selected dimensions can be fewer when class-wise selections overlap.}
\label{tab:selected_feature_protocol}
\begin{tabular}{rrrrrr}
\toprule
$k$ per class &
\begin{tabular}[c]{@{}c@{}}SAE/random\\features\end{tabular} &
\begin{tabular}[c]{@{}c@{}}Raw\\features\end{tabular} &
\begin{tabular}[c]{@{}c@{}}SAE\\selected\end{tabular} &
\begin{tabular}[c]{@{}c@{}}Raw\\selected\end{tabular} &
\begin{tabular}[c]{@{}c@{}}Random SAE\\subset\end{tabular} \\
\midrule
1 & 9 & 9 & 0.858 $\pm$ 0.003 & 0.873 $\pm$ 0.002 & 0.300 $\pm$ 0.111 \\
3 & 27 & 27 & 0.921 $\pm$ 0.002 & 0.927 $\pm$ 0.002 & 0.615 $\pm$ 0.084 \\
5 & 45 & 45 & 0.925 $\pm$ 0.001 & 0.932 $\pm$ 0.002 & 0.738 $\pm$ 0.064 \\
10 & 90 & 89 & 0.936 $\pm$ 0.001 & 0.936 $\pm$ 0.002 & 0.875 $\pm$ 0.029 \\
\bottomrule
\end{tabular}
\end{table}

\FloatBarrier
\section{Additional Qualitative Candidate-Concept Examples}\label{app:qualitative}

\begin{figure}[H]
\centering
\includegraphics[width=0.8\textwidth]{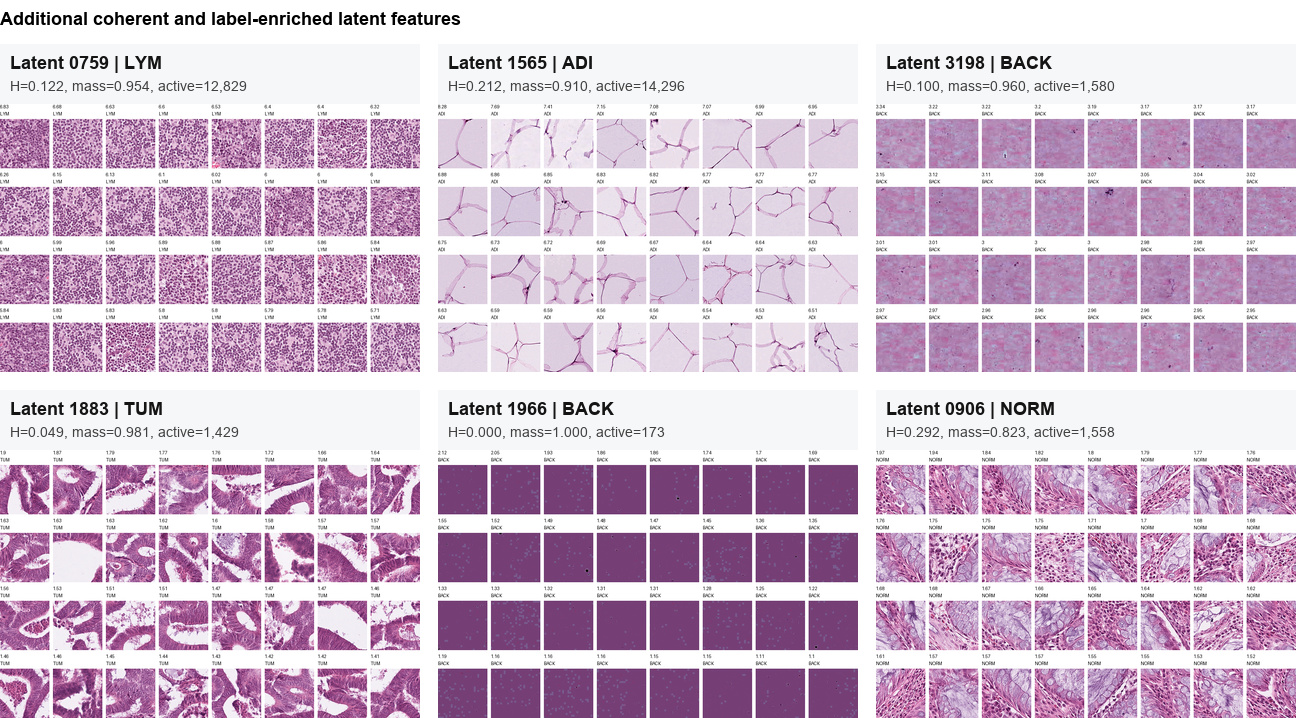}
\caption{\textbf{Additional coherent and label-enriched latent features.} Each panel shows top-activating NCT-CRC patches for one active latent feature, together with compact feature-level statistics from the selected BatchTopK SAE.}
\label{fig:candidate_concepts_coherent}
\end{figure}

\begin{figure}[H]
\centering
\includegraphics[width=0.8\textwidth]{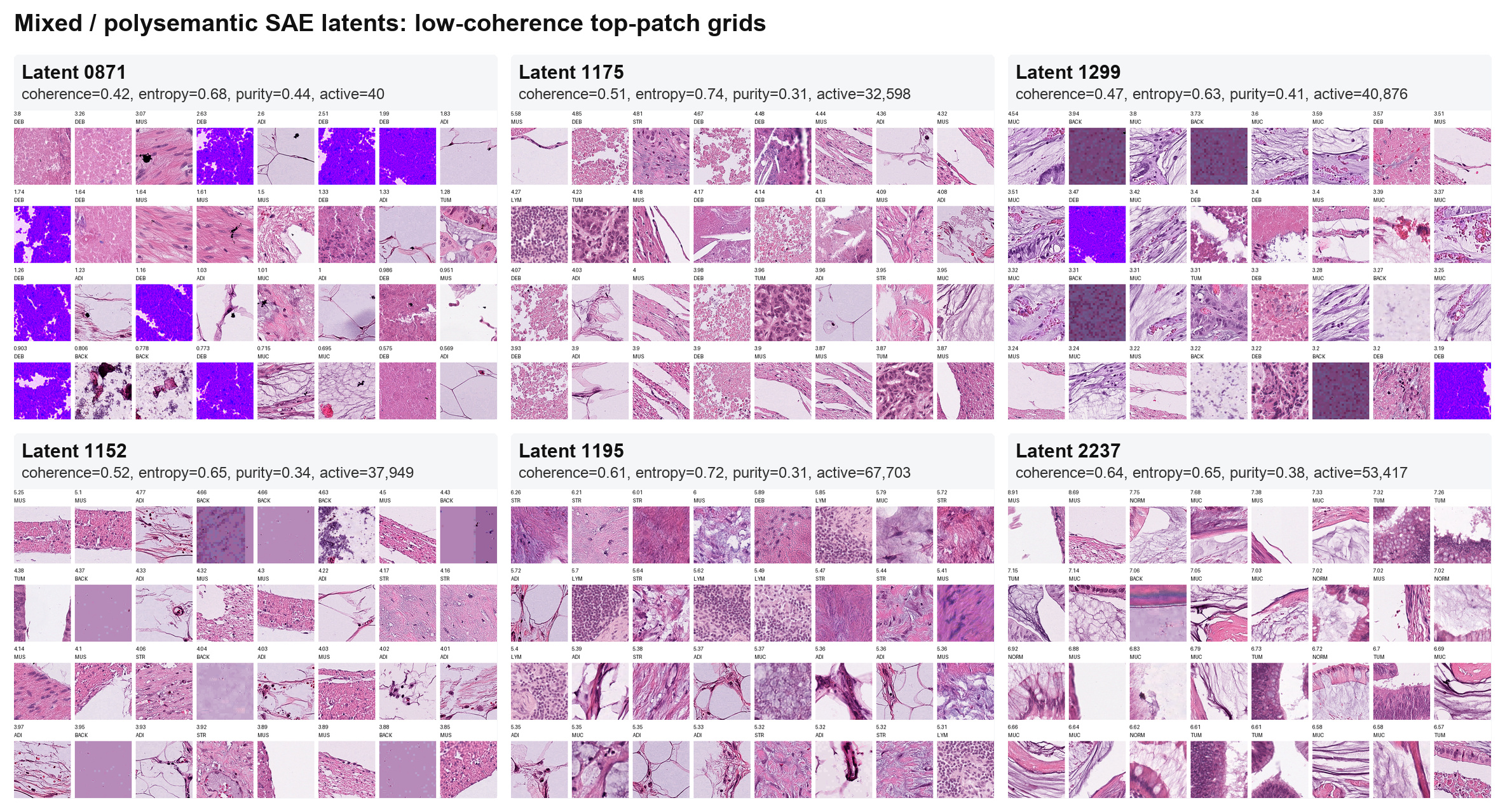}
\caption{\textbf{Mixed, ambiguous, and artifact-associated latent features.} These examples illustrate broad, weakly selective, or technically driven activation behavior and motivate treating SAE latents as candidate concepts requiring expert assessment.}
\label{fig:candidate_concepts_mixed}
\end{figure}

This section provides additional qualitative evidence for candidate-concept assessment. Figure~\ref{fig:candidate_concepts_coherent} presents further examples of coherent and label-enriched latent features, while Figure~\ref{fig:candidate_concepts_mixed} highlights mixed, ambiguous, and artifact-associated features. Together, these examples illustrate the diversity of activation behaviors observed in the learned sparse dictionary and motivate treating SAE latents as candidate concepts requiring expert assessment.

\subsection{WSI Inspection Interface}\label{app:wsi}
\begin{figure}[H]
\centering
\includegraphics[width=0.9\textwidth]{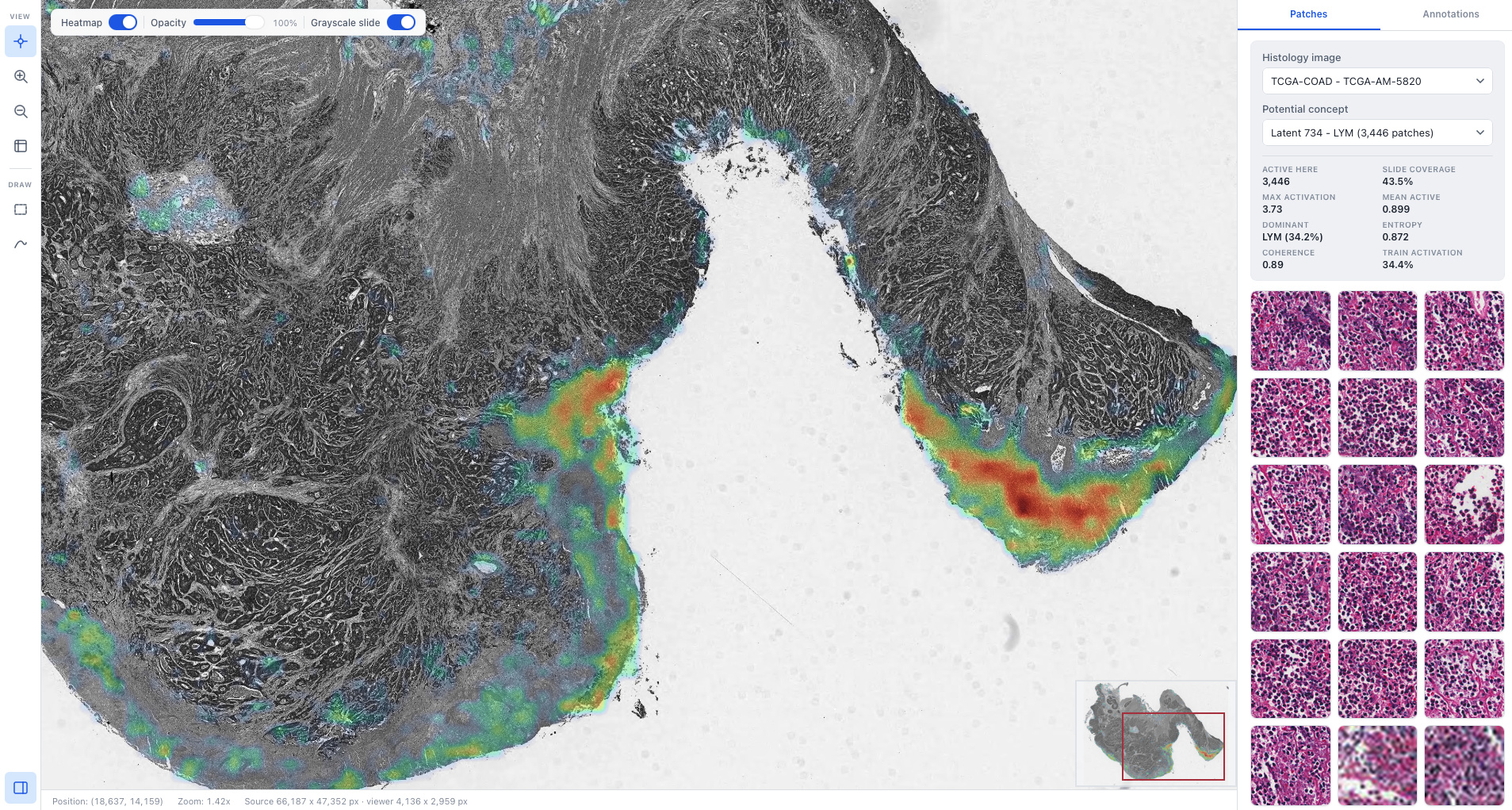}
\caption{\textbf{Screenshot of the WSI-level latent inspection interface.}
Example view for one selected SAE latent feature on a TCGA-COAD whole-slide image. The interface links top-activating patches with the corresponding slide-level activation overlay, supporting qualitative inspection of spatial deployment within larger tissue context.}
\label{fig:tcga_wsi_examples}
\end{figure}

To support expert inspection, we developed an interactive interface that links SAE latent features, top-activating patches, and WSI-level activation overlays. This allows candidate concepts to be examined within larger tissue context rather than through isolated patch retrieval alone. Figure~\ref{fig:tcga_wsi_examples} shows an example interface view used during qualitative assessment of latent-feature deployment patterns. Table~\ref{tab:tcga_wsi_protocol} lists the WSI processing parameters.

\begin{table}[H]
\centering
\caption{\textbf{TCGA-COAD WSI qualitative overlays.} These WSI overlays are used for spatial inspection of candidate concepts.}
\label{tab:tcga_wsi_protocol}
\begin{tabular}{lc}
\toprule
Parameter & Value \\
\midrule
WSI source & 5 TCGA-COAD SVS slides \\
Selected SAE & BatchTopK, $m=4096$, target $L_0=64$, seed 1 \\
Embedding call & CONCH image encoder, unnormalized output \\
Analysis resolution & $\approx 0.5\,\mu\mathrm{m}$/px \\
Encoded patch size & $224 \times 224$ pixels \\
Native WSI footprint / stride & 444--456 px / 444--456 px \\
Tissue filter & white-threshold tissue mask \\
Minimum tissue fraction & 0.20 \\
Retained tissue patches & 49,125 total; 1,419--18,564 per slide \\
Rendered latents & Up to 96 active latents per slide \\
Overlay style & contiguous patch-cell heatmap \\
\bottomrule
\end{tabular}
\end{table}

\FloatBarrier
\section{UNI2-h Robustness Details}\label{app:uni2}

\begin{figure}[H]
\centering
\includegraphics[width=0.96\textwidth]{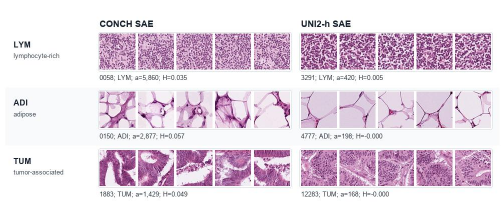}
\caption{\textbf{Candidate features of similar type in independently trained CONCH and UNI2-h SAEs.} Top-activating NCT-CRC patches for features associated with lymphocytes, adipose tissue and tumor epithelium. Visual similarity does not establish a one-to-one correspondence between features.}
\label{fig:shared_candidates}
\end{figure}

UNI2-h SAEs use the same architecture, optimizer, learning rate ($10^{-3}$), batch size (4096) and early stopping as the CONCH SAEs, with input dimension 1536, dictionary width 12288 and target $L_0=192$. This matches the CONCH configuration in expansion factor (8$\times$) and target activation fraction ($L_0/m=1.5625\%$). Reconstruction statistics in Table~\ref{tab:backbone_reconstruction} are averaged over seeds 0--4; feature-level analyses use seed 3. On the combined train and validation split, 10,930 of 12,288 features activate at least once, and 6,118 have at least 32 positive responses and enter the top-32 retrieval analysis. Table~\ref{tab:uni2h_controls} compares these features with signed directions using the best-tail setting of Appendix~\ref{app:tails}; Figure~\ref{fig:best_tail_both} shows the same comparison next to the CONCH values from Table~\ref{tab:tail_handling}, and Figure~\ref{fig:shared_candidates} shows candidate features of similar type found independently in the CONCH and UNI2-h SAEs.

\begin{table}[H]
\centering
\small
\caption{\textbf{UNI2-h top-32 retrieval compared with signed directions.} Each raw dimension (1,536), PCA component (512) and random direction (512) uses its more coherent tail; entropy is measured on the same patch set.}
\label{tab:uni2h_controls}
\begin{tabular}{lcc}
\toprule
Method & Coherence $\uparrow$ & Entropy $\downarrow$ \\
\midrule
SAE feature & 0.605 & 0.101 \\
Raw UNI2-h dim & 0.497 & 0.232 \\
Random direction & 0.485 & 0.249 \\
PCA component & 0.290 & 0.571 \\
\bottomrule
\end{tabular}
\end{table}

\begin{figure}[H]
\centering
\includegraphics[width=0.8\textwidth]{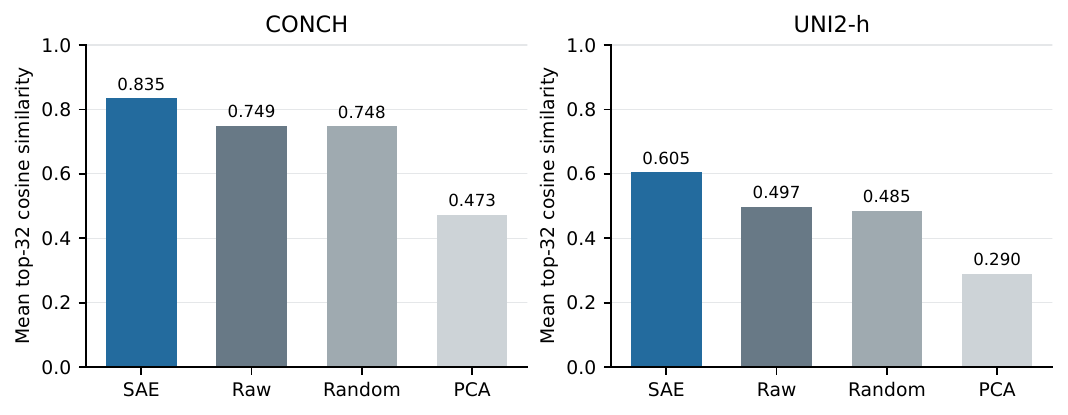}
\caption{\textbf{Top-32 retrieval coherence under the best-tail setting for CONCH (left) and UNI2-h (right).} For every signed raw, random or PCA direction, the more coherent tail is used; SAE features use their highest positive activations. Each panel is a comparison within one embedding space, so absolute values should not be compared across models.}
\label{fig:best_tail_both}
\end{figure}

\FloatBarrier
\end{document}